\documentclass[10pt,journal,compsoc]{IEEEtran}
\usepackage{subfigure}
\usepackage{graphicx}
\usepackage{caption} 
\usepackage{booktabs}
\usepackage{amsfonts,amssymb, amsthm, amsmath}
\usepackage{rotating}
\usepackage{multirow, multicol}
\usepackage{makecell}
\usepackage{diagbox}
\usepackage{soul}
\usepackage[linesnumbered,ruled, vlined]{algorithm2e}
\usepackage{algpseudocode}
\usepackage{arydshln} 
\usepackage[usenames,dvipsnames]{xcolor} 
\usepackage[pagebackref,breaklinks,colorlinks]{hyperref}
\usepackage[capitalize]{cleveref}
\usepackage{caption}
\usepackage[super]{nth}
\usepackage{microtype}
\usepackage[most]{tcolorbox}
\usepackage{array}
\usepackage{mdframed}
\usepackage{enumitem}
\usepackage{mathtools}
\usepackage{float}

\crefname{section}{Sec.}{Secs.}
\Crefname{section}{Section}{Sections}
\Crefname{table}{Table}{Tables}
\crefname{table}{Tab.}{Tabs.}

\ifCLASSOPTIONcompsoc
\usepackage[nocompress]{cite}
\else
\usepackage{cite}
\fi
\ifCLASSINFOpdf
\else
\fi
\begin{document}
		%
		\title{Graph-Augmented Reasoning:\\ Evolving Step-by-Step Knowledge Graph Retrieval for LLM Reasoning}
		%
		%
		%
		%
		
		\author{Wenjie Wu, Yongcheng Jing, Yingjie Wang, Wenbin Hu,  Dacheng Tao
			\IEEEcompsocitemizethanks{
				\IEEEcompsocthanksitem Wenjie Wu and Wenbin Hu are with Wuhan University (Email: hwb@whu.edu.cn).
				\IEEEcompsocthanksitem Yongcheng Jing, Yingjie Wang, and Dacheng Tao are with Nanyang Technological University (Email: dacheng.tao@ntu.edu.sg).
			}
		}

	\IEEEtitleabstractindextext{%
		\begin{abstract}
			Recent large language model (LLM) reasoning, despite its success, suffers from limited domain knowledge, susceptibility to hallucinations, and constrained reasoning depth, particularly in small-scale models deployed in resource-constrained environments. This paper presents the first investigation into integrating step-wise knowledge graph retrieval with step-wise reasoning to address these challenges, introducing a novel paradigm termed as graph-augmented reasoning. Our goal is to enable frozen, small-scale LLMs to retrieve and process relevant mathematical knowledge in a step-wise manner, enhancing their problem-solving abilities without additional training. To this end, we propose KG-RAR, a framework centered on process-oriented knowledge graph construction, a hierarchical retrieval strategy, and a universal post-retrieval processing and reward model (PRP-RM) that refines retrieved information and evaluates each reasoning step. 
Experiments on the Math500 and GSM8K benchmarks across six models demonstrate that KG-RAR yields encouraging results, achieving a 20.73\% relative improvement with Llama-3B on Math500.
		\end{abstract}
		
		}

	\maketitle

	\IEEEdisplaynontitleabstractindextext

	%
	\IEEEpeerreviewmaketitle

	\IEEEraisesectionheading{
		\section{Introduction}
		\label{sec:intro}
	}
        \IEEEPARstart{E}{nhancing} the reasoning capabilities of large language models (LLMs) continues to be a major challenge \cite{huang2022towards, sun2023survey, plaat2024reasoning}. Conventional methods, such as chain-of-thought (CoT) prompting \cite{wei2022chain}, improve inference by encouraging step-by-step articulation \cite{wang2022self, zhou2022docprompting, kojima2022large, creswell2022selection, reflexion, yao2024tree}, while external tool usage and domain-specific fine-tuning further refine specific task performance \cite{chen2022program, yamauchi2023lpml, zhuang2023toolqa, roziere2023code, yu2023metamath, wu2024mathchat}. Most recently, \emph{o1-like multi-step reasoning} has emerged as a paradigm shift \cite{wu2024comparative, zhang2024llama, zhang2024o1, wang2024drt, wang2024openr, luo2025o1}, leveraging \emph{test-time compute} strategies \cite{wang2022self, feng2023alphazero, hao2023reasoning, snell2024scaling, chen2024simple}, exemplified by reasoning models like GPT-o1 \cite{o1} and DeepSeek-R1 \cite{deepseekai2025deepseekr1incentivizingreasoningcapability}. These approaches, including Best-of-N \cite{lightman2023let} and Monte Carlo Tree Search \cite{feng2023alphazero} , allocate additional computational resources during inference to dynamically refine reasoning paths \cite{cobbe2021gsm8k, uesato2022solving, lightman2023let, snell2024scaling}.

\begin{figure}[t]
\vskip 0.2in
\begin{center}
\centerline{\includegraphics[width=\columnwidth]{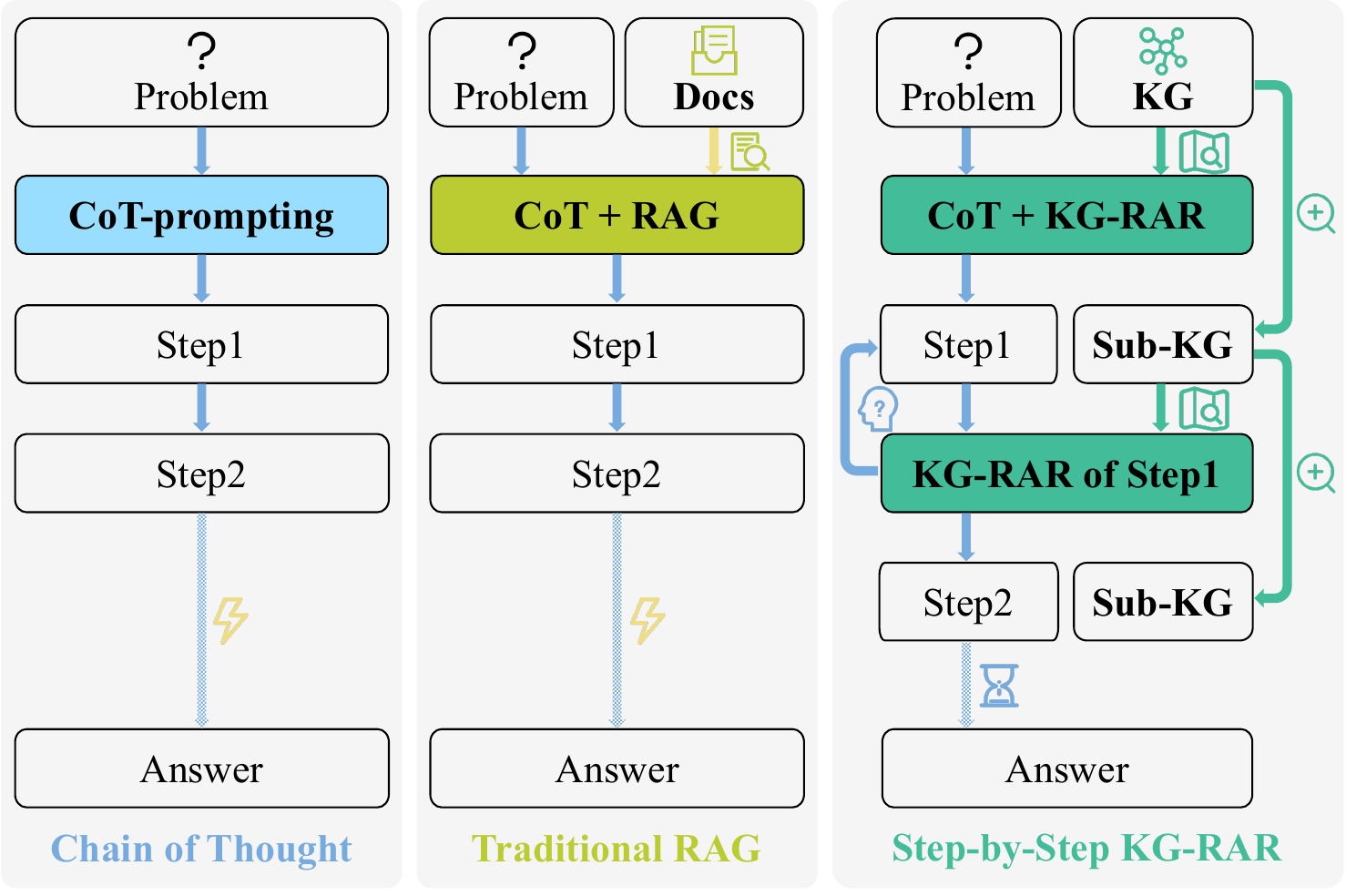}}
\caption{Illustration of the proposed step-by-step knowledge graph retrieval for o1-like reasoning, which dynamically retrieves and utilises structured sub-graphs (Sub-KGs) during reasoning. Our approach iteratively refines the reasoning process by retrieving relevant Sub-KGs at each step, enhancing accuracy, consistency, and reasoning depth for complex tasks, thereby offering a novel form of scaling test-time computation.}
\label{fig:step-by-step-kg-rag}
\end{center}
\vskip -0.2in
\end{figure}

Despite encouraging advancements in o1-like reasoning, LLMs—particularly smaller and less powerful variants—continue to struggle with complex reasoning tasks in mathematics and science \cite{sun2023survey, plaat2024reasoning, satpute2024can, mirzadeh2024gsm}. These challenges arise from \emph{insufficient domain knowledge}, \emph{susceptibility to hallucinations}, and \emph{constrained reasoning depth} \cite{huang2023survey, banerjee2024llms, xu2024hallucination}. 
Given the novelty of o1-like reasoning, effective solutions to these issues remain largely unexplored, with few studies addressing this gap in the literature \cite{wu2024comparative, zhang2024llama, luo2025o1}.
One potential solution from the pre-o1 era is \emph{retrieval-augmented generation (RAG)}, which has been shown to mitigate hallucinations and factual inaccuracies by retrieving relevant information from external knowledge sources (Fig.~\ref{fig:step-by-step-kg-rag}, the \nth{2} column) \cite{gao2023retrieval, fan2024survey, peng2024graph}. However, in the context of o1-like multi-step reasoning, traditional RAG faces two significant challenges: 
\begin{itemize}[leftmargin=*, itemsep=0pt, parsep=0pt, topsep=0pt, partopsep=0pt]
\item {\emph{ \textbf{(1) Step-wise hallucinations}}: {LLMs may hallucinate during \emph{intermediate steps}—a problem not addressed by applying RAG solely to the initial prompt \cite{barnett2024seven, wang2024rat};}
\item \emph{ \textbf{(2) Missing structured relationships}}: {Traditional RAG may retrieve information that lacks the \emph{structured relationships} necessary for complex reasoning tasks, leading to inadequate augmentation that fails to capture the depth required for accurate reasoning \cite{peng2024graph, hu2024grag}.}}
\end{itemize}

In this paper, we strive to address both challenges by introducing a novel \emph{graph-augmented multi-step reasoning} scheme to enhance LLMs' o1-like reasoning capability, as depicted in Fig.~\ref{fig:step-by-step-kg-rag}. Our idea is motivated by the recent success of knowledge graphs (KGs) in knowledge-based question answering and fact-checking \cite{li2023chain, he2024g, chang2024communitykg, mu2024predicting, luo2023reasoning, sun2023think, liu2024knowledge}. 
Recent advances have demonstrated the effectiveness of KGs in augmenting prompts with retrieved knowledge or enabling LLMs to query KGs for factual information \cite{luo2024graph, zhang2024knowgpt}. However, little attention has been given to improving step-by-step reasoning for complex tasks with KGs \cite{choudhary2023complex, zhao2024stepwise}, such as mathematical reasoning, which requires iterative logical inference rather than simple knowledge retrieval.

To fill this gap, the objective of the proposed graph-augmented reasoning paradigm is to integrate {\emph{structured KG retrieval}} into the reasoning process in a step-by-step manner, providing contextually relevant information {\emph{at each reasoning step to refine reasoning paths and mitigate step-wise inaccuracies and hallucinations}}, thereby addressing both aforementioned challenges simultaneously. This approach operates without additional training, making it particularly well-suited for small-scale LLMs in resource-constrained environments. Moreover, it extends test-time compute by incorporating external knowledge into the reasoning context, transitioning from direct CoT to step-wise guided retrieval and reasoning.

Nevertheless, implementing this graph-augmented reasoning paradigm is accompanied with several key issues: (1) Frozen LLMs struggle to query KGs effectively \cite{luo2024graph}, necessitating a dynamic integration strategy for iterative incorporation of graph-based knowledge; (2) Existing KGs primarily encode static facts rather than the procedural knowledge required for multi-step reasoning \cite{ji2021survey, wang2022math}, highlighting the need for process-oriented KGs; (3) Reward models, which are essential for validating reasoning steps \cite{cobbe2021gsm8k, lightman2023let}, often require costly fine-tuning and suffer from poor generalization \cite{zheng2024processbench}, underscoring the need for a universal, training-free scoring mechanism tailored to KG.

To address these issues, we propose KG-RAR, a step-by-step knowledge graph based retrieval-augmented reasoning framework that retrieves, refines, and reasons using structured knowledge graphs in a step-wise manner.
Specifically, to enable effective KG querying, we design a hierarchical retrieval strategy in which questions and reasoning steps are progressively matched to relevant subgraphs, dynamically narrowing the search space.
Also, we present a process-oriented math knowledge graph (MKG) construction method that encodes step-by-step procedural knowledge, ensuring that LLMs retrieve and apply structured reasoning sequences rather than static facts. Furthermore, we introduce the post-retrieval processing and reward model (PRP-RM)—a training-free scoring mechanism that refines retrieved knowledge before reasoning and evaluates step correctness in real time.
By integrating structured retrieval with test-time computation, our approach mitigates reasoning inconsistencies, reduces hallucinations, and enhances stepwise verification—all without  additional training.

In sum, our contribution is therefore the first attempt that dynamically integrates step-by-step KG retrieval into an o1-like multi-step reasoning process. This is achieved through our proposed hierarchical retrieval, process-oriented graph construction method, and PRP-RM—a training-free scoring mechanism that ensures retrieval relevance and step correctness. Experiments on Math500 and GSM8K validate the effectiveness of our approach across six smaller models from the Llama3 and Qwen2.5 series. The best-performing model, Llama-3B on Math500, achieves a 20.73\% relative improvement over CoT prompting, followed by Llama-8B on Math500 with a 15.22\% relative gain and Llama-8B on GSM8K with an 8.68\% improvement.
	
	\section{Related Work}
	\label{sec:related work}

            \begin{figure*}[t]
\begin{center}
\centerline{\includegraphics[width=0.95\linewidth]{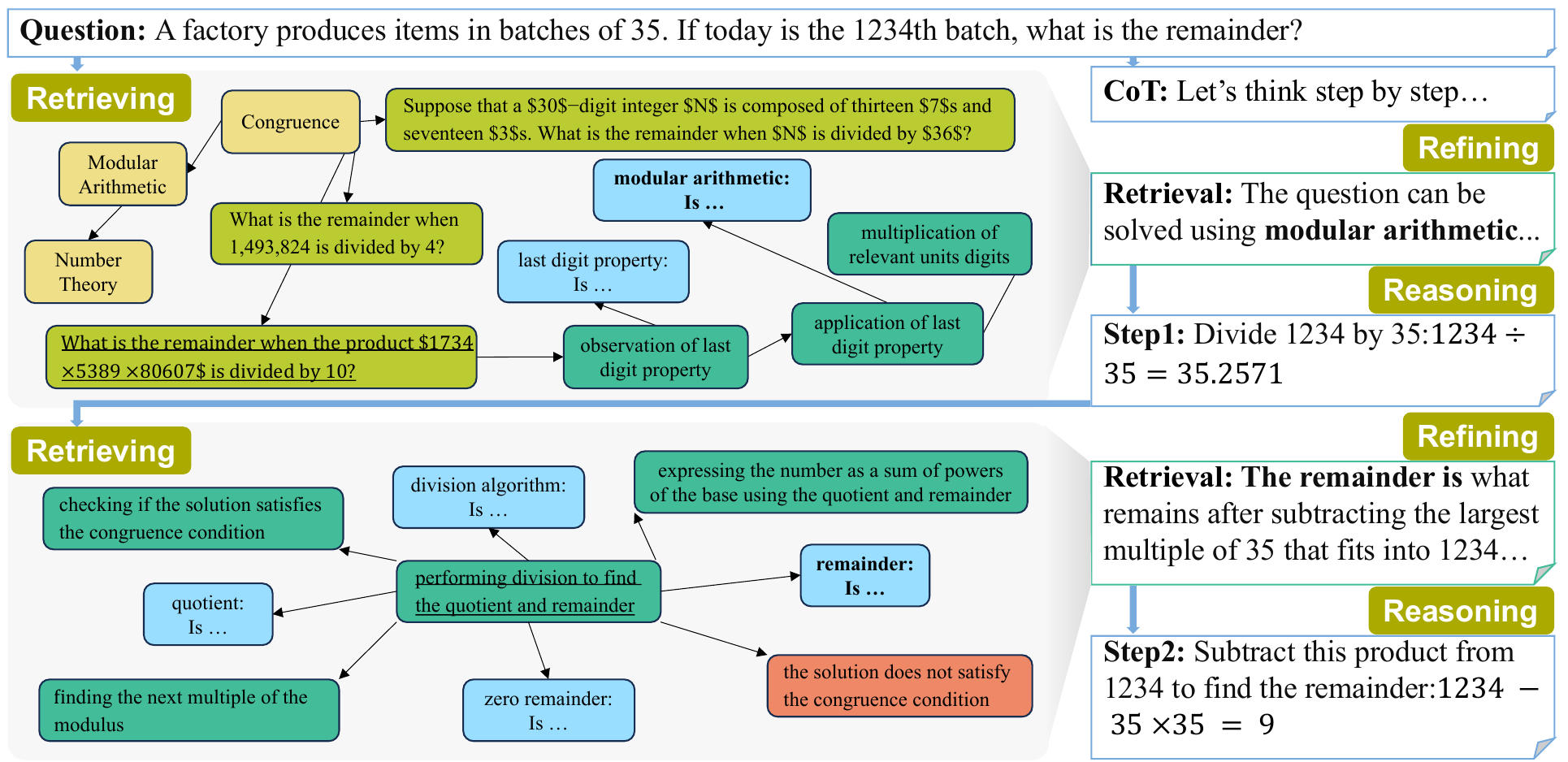}}
\caption{Example of Step-by-Step KG-RAR's iterative process: 1) \textbf{Retrieving:} For a given question or intermediate reasoning step, the KG is retrieved to find the most similar problem or procedure (underlined in the figure) and extract its subgraph as the raw retrieval. 2) \textbf{Refining:} A frozen LLM processes the raw retrieval to generate a refined and targeted context for reasoning. 3) \textbf{Reasoning:} Using the refined retrieval, another LLM reflects on previous steps and generates next intermediate reasoning steps. This iterative workflow refines and guides the reasoning path to problem-solving.}
\label{fig:overview}
\end{center}
\vskip -0.2in
\end{figure*}
    
	\subsection{LLM Reasoning}
	Large Language Models (LLMs) have advanced in structured reasoning through techniques like Chain-of-Thought (CoT) \cite{wei2022chain}, Self-Consistency \cite{wang2022self}, and Tree-of-Thought \cite{yao2024tree}, improving inference by generating intermediate steps rather than relying on greedy decoding \cite{zhou2022docprompting, kojima2022large, creswell2022selection, graphofthought, zelikman2022star, reflexion}. Recently, GPT-o1-like reasoning has emerged as a paradigm shift \cite{wu2024comparative, zhang2024llama, snell2024scaling, luo2025o1}, leveraging Test-Time Compute strategies such as Best-of-N \cite{lightman2023let}, Beam Search \cite{wang2022self}, and Monte Carlo Tree Search \cite{feng2023alphazero}, often integrated with reward models to refine reasoning paths dynamically \cite{cobbe2021gsm8k, uesato2022solving, lightman2023let, li2022making, yu2024ovm, zhang2024generative}. Reasoning models like DeepSeek-R1 exemplify this trend by iteratively searching, verifying, and refining solutions, significantly enhancing inference accuracy and robustness \cite{o1, deepseekai2025deepseekr1incentivizingreasoningcapability}. However, these methods remain computationally expensive and challenging for small-scale LLMs, which struggle with hallucinations and inconsistencies due to limited reasoning capacity and lack of domain knowledge \cite{satpute2024can, huang2023survey, plaat2024reasoning}.

	\subsection{Knowledge Graphs Enhanced LLM Reasoning}
	Knowledge Graphs (KGs) are structured repositories of interconnected entities and relationships, offering efficient graph-based knowledge representation and retrieval \cite{paulheim2016knowledge, wang2017knowledge,jing2021meta,jing2023deep, ji2021survey}. Prior work integrating KGs with LLMs has primarily focused on knowledge-based reasoning tasks such as knowledge-based question answering \cite{li2023chain, he2024g, sanmartin2024kg, wang2024knowledge}, fact-checking \cite{chang2024communitykg, mu2024predicting, kau2024combining}, and entity-centric reasoning \cite{jiang2023structgpt, luo2023reasoning, chai2023graphllm, sun2023think, liu2024knowledge}. However, in these tasks, "reasoning" is predominantly limited to identifying and retrieving static knowledge rather than performing iterative, multi-step logical computations \cite{zhu2024llms, pan2024unifying, agrawal2023can}. In contrast, our work is to integrate KGs with LLMs for o1-like reasoning in domains such as mathematics, where solving problems demands dynamic, step-by-step inference rather than static knowledge retrieval.

	\subsection{Reward Models}
    Reward models are essential across various domains such as computer vision \cite{pinto2023tuning,jing2021turning}. Notably, they play a crucial role in aligning LLM outputs with human preferences by evaluating accuracy, relevance, and logical consistency \cite{kwon2023reward, cao2024survey, wang2024comprehensive}. Fine-tuned reward models, including Outcome Reward Models (ORMs) \cite{cobbe2021gsm8k} and Process Reward Models (PRMs) \cite{uesato2022solving, lightman2023let, yu2024ovm}, improve validation accuracy but come at a high training cost and often lack generalization across diverse tasks \cite{zheng2024processbench}. Generative reward models \cite{zhang2024generative} further enhance performance by integrating CoT reasoning into reward assessments, leveraging Test-Time Compute to refine evaluation. However, the reliance on fine-tuning makes these models resource-intensive and limits adaptability \cite{zheng2024processbench}. This underscores the need for universal, training-free scoring mechanisms that maintain robust performance while ensuring computational efficiency across various reasoning domains.

        \section{Pre-analysis}
        \label{sec: pre-analysis}
        \subsection{Motivation and Problem Definition}

\textbf{Motivation.} LLMs have demonstrated remarkable capabilities across various domains \cite{brown2020language, wei2022chain, react, least-to-most,wang2024chain}, yet their proficiency in complex reasoning tasks remains limited \cite{plaat2024reasoning, zhang2024llm, yu2024natural}. Challenges such as hallucinations \cite{huang2023survey, huang2024survey}, inaccuracies, and difficulties in handling complex, multi-step reasoning due to insufficient reasoning depth are particularly evident in smaller models or resource-constrained environments \cite{chu2023survey, ahn2024large, li2024llms}. Moreover, traditional reward models, including ORMs \cite{cobbe2021gsm8k} and PRMs \cite{uesato2022solving, lightman2023let}, require extensive fine-tuning, incurring significant computational costs for dataset collection, GPU usage, and prolonged training time \cite{wang2023aligning, shen2023large, balseiro2024survey}. Despite these efforts, fine-tuned reward models often suffer from poor generalization, restricting their effectiveness across diverse reasoning tasks \cite{zheng2024processbench}.

To simultaneously overcome these challenges, this paper introduces a novel paradigm tailored for o1-like multi-step reasoning:

\textbf{Remark 3.1 (Graph-Augmented Multi-Step Reasoning).} \textit{The goal of graph-augmented reasoning is to enhance the step-by-step reasoning ability of frozen LLMs by integrating external knowledge graphs (KGs), eliminating the need for additional fine-tuning.}

The proposed graph-augmented scheme aims to offer the following unique advantages:
\begin{itemize}[leftmargin=*, itemsep=0pt, parsep=0pt, topsep=0pt, partopsep=0pt]
    \item \textbf{Improving Multi-Step Reasoning:} Enhances reasoning capabilities, particularly for small-scale LLMs in resource-constrained environments;
    \item \textbf{Scaling Test-Time Compute:} Introduces a novel dimension of scaling test-time compute through dynamic integration of external knowledge;
    \item \textbf{Transferability Across Reasoning Tasks:} By leveraging domain-specific KGs, the framework can be easily adapted to various reasoning tasks, enabling transferability across different domains.
\end{itemize}

\subsection{Challenge Analysis}

However, implementing the proposed graph-augmented reasoning paradigm presents several critical challenges:
\begin{itemize}[leftmargin=*, itemsep=0pt, parsep=0pt, topsep=0pt, partopsep=0pt]
    \item \textbf{Effective Integration:} How can KGs be efficiently integrated with LLMs to support step-by-step reasoning without requiring model modifications? Frozen LLMs cannot directly query KGs effectively \cite{luo2024graph}. Additionally, since LLMs may suffer from hallucinations and inaccuracies during intermediate reasoning steps \cite{huang2023survey, huang2024survey, wang2024rat}, it is crucial to dynamically integrate KGs at each step rather than relying solely on static knowledge retrieved at the initial stage;
    \item \textbf{Knowledge Graph Construction:} How can we design and construct process-oriented KGs tailored for LLM-driven multi-step reasoning? Existing KGs predominantly store static knowledge rather than the procedural and logical information required for reasoning \cite{ji2021survey, wang2022math, tomaszuk2023mmlkg}. A well-structured KG that represents reasoning steps, dependencies, and logical flows is necessary to support iterative reasoning;
    \item \textbf{Universal Scoring Mechanism:} How can we develop a training-free reward mechanism capable of universally evaluating reasoning paths across diverse tasks without domain-specific fine-tuning? Current approaches depend on fine-tuned reward models, which are computationally expensive and lack adaptability \cite{zheng2024processbench}. A universal, training-free scoring mechanism leveraging frozen LLMs is essential for scalable and efficient reasoning evaluation.
\end{itemize}

To address these challenges and unlock the full potential of graph-augmented reasoning, we propose a \emph{Step-by-Step Knowledge Graph based Retrieval-Augmented Reasoning (KG-RAR)} framework, accompanied by a dedicated \emph{Post-Retrieval Processing and Reward Model (PRP-RM)}, which will be elaborated in the following section.

	\section{Proposed Approach}
	\label{sec: proposed approach}	
	\subsection{Overview}

Our objective is to integrate KGs for o1-like reasoning with frozen, small-scale LLMs in a training-free and universal manner. This is achieved by integrating a step-by-step knowledge graph based retrieval-augmented reasoning (KG-RAR) module within a structured, iterative reasoning framework. As shown in Figure~\ref{fig:overview}, the iterative process comprises three core phases: retrieving, refining, and reasoning.

	\subsection{Process-Oriented Math Knowledge Graph}

    \begin{figure}[ht]
\begin{center}
\centerline{\includegraphics[width=\columnwidth]{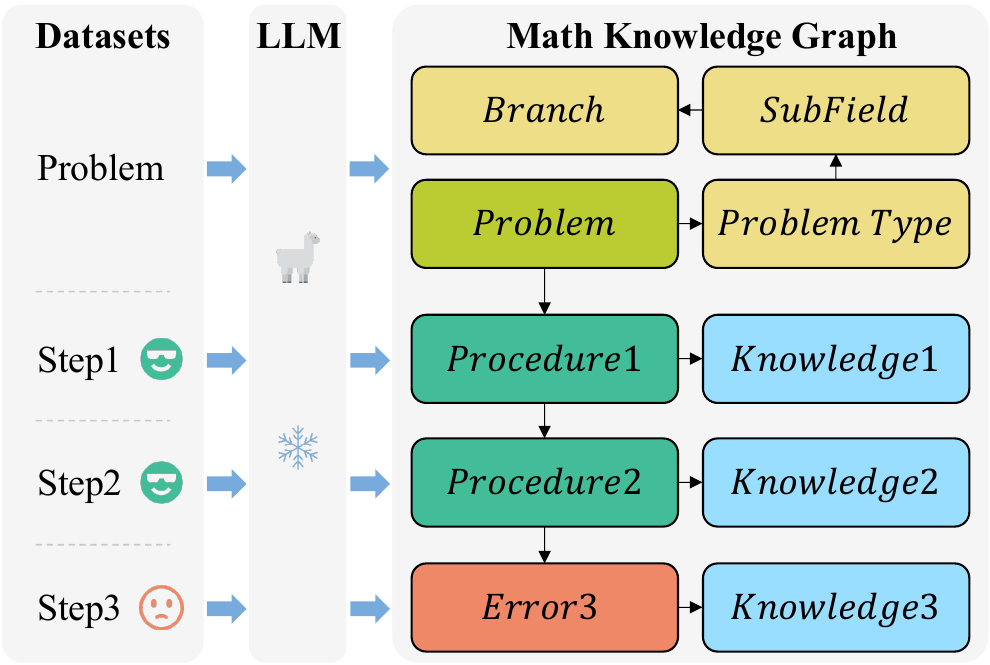}}
\caption{Pipeline for constructing the process-oriented math knowledge graph from process supervision datasets.}
\label{fig:pic_mkg_building}
\end{center}
\vskip -0.2in
\end{figure}

To support o1-like multi-step reasoning, we construct a Mathematical Knowledge Graph tailored for multi-step logical inference. Public process supervision datasets, such as PRM800K \cite{lightman2023let}, provide structured problem-solving steps annotated with artificial ratings. Each sample will be decomposed into the following structured components: \textit{branch, subfield, problem type, problem, procedures, errors, and related knowledge}.

The Knowledge Graph is formally defined as: $G = (V, E)$, where \(V\) represents nodes—including problems, procedures, errors, and mathematical knowledge—and \(E\) represents edges encoding their relationships (e.g., "derived from," "related to").

As shown in Figure~\ref{fig:pic_mkg_building}, for a given problem \( P \) with solutions \( S_1, S_2, \dots, S_n \) and human ratings, the structured representation is:
\begin{equation*}
    P \mapsto \{B_p, F_p, T_p, \mathbf{r}\},
    \vspace{-1em}
\end{equation*}
\begin{align*}
    S_i^\text{good} &\mapsto \{S_i, K_i, \mathbf{r}_i^\text{good}\}, \quad
    S_i^\text{bad} \mapsto \{E_i, K_i, \mathbf{r}_i^\text{bad}\},
\end{align*}
where \(B_p, F_p, T_p\) represent the branch, subfield, and type of \(P\), respectively. The symbols \(S_i\) and \(E_i\) denote the procedures derived from correct steps and the errors from incorrect steps, respectively. Additionally, \(K_i\) contains relevant mathematical knowledge. The relationships between problems, steps, and knowledge are encoded through \(\mathbf{r}, \mathbf{r}_i^\text{good}, \mathbf{r}_i^\text{bad}\), which capture the edge relationships linking these elements.

To ensure a balance between computational efficiency and quality of KG, we employ a Llama-3.1-8B-Instruct model to process about 10,000 unduplicated samples from PRM800K. The LLM is prompted to output structured JSON data, which is subsequently transformed into a Neo4j-based MKG. This process yields a graph with approximately \(80,000\) nodes and \(200,000\) edges, optimized for efficient retrieval.

\subsection{Step-by-Step Knowledge Graph Retrieval}

\noindent\textbf{KG-RAR for Problem Retrieval.} For a given test problem \( Q \), the most relevant problem \( P^* \in V_p \) and its subgraph are retrieved to assist reasoning. The retrieval pipeline comprises the following steps:

1. Initial Filtering: Classify \( Q \) by \(B_q, F_q, T_q\) (branch, subfield, and problem type). The candidate set \(V_Q \subset V_p\) is filtered hierarchically, starting from \(T_q\), expanding to \(F_q\), and then to \(B_q\) if no exact match is found.

2. Semantic Similarity Scoring:
\[
P^* = \arg\max_{P \in V_Q} \cos(\mathbf{e}_Q, \mathbf{e}_P),
    \vspace{-1em}
\]
where:
\[
\cos(\mathbf{e}_Q, \mathbf{e}_P) = \frac{\langle \mathbf{e}_Q, \mathbf{e}_P \rangle}{\|\mathbf{e}_Q\| \|\mathbf{e}_P\|}
\]
and \(\mathbf{e}_Q, \mathbf{e}_P \in \mathbb{R}^d\) are embeddings of \(Q\) and \(P\), respectively.

3. Context Retrieval: Perform Depth-First Search (DFS) on \(G\) to retrieve procedures (\(S_p\)), errors (\(E_p\)), and knowledge (\(K_p\)) connected to \(P^*\).

\begin{algorithm}[t]
\caption{KG-RAR for Problem Retrieval}
\KwIn{Test problem \(Q\) and MKG \(G\)}
\KwOut{Most relevant problem \(P^*\) and its context \((S_p, E_p, K_p)\)}
\BlankLine
Filter \(G\) using \(B_q\), \(F_q\), and \(T_q\) to obtain \(V_Q\)\;
\BlankLine
\ForEach{\(P \in V_Q\)}{
  Compute \(\text{Sim}_{\text{semantic}}(Q, P)\)\;
}
\BlankLine
\(P^* \gets \arg\max_{P \in V_Q} \text{Sim}_{\text{semantic}}(Q, P)\)\;
\BlankLine
Retrieve \(S_p\), \(E_p\), and \(K_p\) from \(P^*\) using DFS\;
\BlankLine
\Return \(P^*\), \(S_p\), \(E_p\), and \(K_p\)\;
\end{algorithm}

\noindent\textbf{KG-RAR for Step Retrieval.} Given an intermediate reasoning step \( S \), the most relevant step \( S^* \in G \) and its subgraph is retrieved dynamically:

1. Contextual Filtering: Restrict the search space \(V_S\) to the subgraph induced by previously retrieved top-k similar problems \( \{P_1, P_2, \dots, P_k\} \in V_Q\).

2. Step Similarity Scoring:
\[
S^* = \arg\max_{S_i \in V_S} \cos(\mathbf{e}_S, \mathbf{e}_{S_i}).
\]

3. Context Retrieval: Perform Breadth-First Search (BFS) on \(G\) to extract subgraph of \(S^*\), including potential next steps, related knowledge, and error patterns.

\subsection{Post-Retrieval Processing and Reward Model}

\noindent\textbf{Step Verification and End-of-Reasoning Detection.} 
Inspired by previous works \cite{zheng2023judging, zheng2024processbench, li2024generation, zhang2024generative}, we use a frozen LLM to evaluate both step correctness and whether reasoning should terminate. The model is queried with an instruction, producing a binary classification decision:
\[
\textit{Is this step correct (Yes/No)? }  \begin{cases}
    \text{Yes} \xrightarrow[Probability]{Token}p(\text{Yes})\\
    \text{No} \xrightarrow[Probability]{Token}p(\text{No})\\
    \text{Other Tokens.}
\end{cases}
\]
The corresponding confidence score for step verification or reasoning termination is computed as:
\[
Score(S, I) = \frac{\exp(p(\text{Yes} | S, I))}{\exp(p(\text{Yes} | S, I)) + \exp(p(\text{No} | S, I))}.
\]
For step correctness, the instruction \( I \) is \textit{"Is this step correct (Yes/No)?"}, while for reasoning termination, the instruction \( I_E \) is \textit{"Has a final answer been reached (Yes/No)?"}.

\noindent\textbf{Post-Retrieval Processing.}
Post-retrieval processing is a crucial component of the retrieval-augmented generation (RAG) framework, ensuring that retrieved information is improved to maximize relevance while minimizing noise \cite{gao2023retrieval, shi2024enhancing, cao2024lego}. 

For a problem \( P \) or a reasoning step \( S \):
\[
\mathcal{R}' = \operatorname{LLM}_{\text{refine}}(P + \mathcal{R} \text{ or } S + \mathcal{R}),
\]
where \( \mathcal{R} \) is the raw retrieved context, and \( \mathcal{R}' \) represents its rewritten, targeted form.

\begin{figure}[t]
\begin{center}
\centerline{\includegraphics[width=\columnwidth]{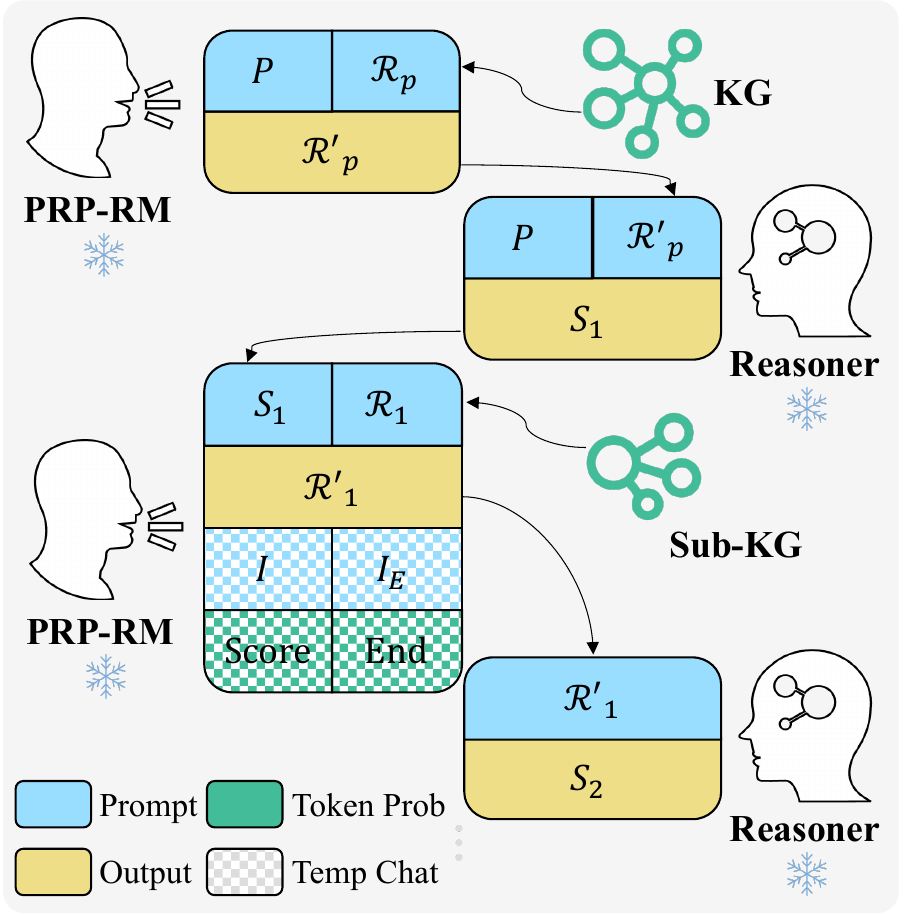}}
\caption{Illustration of the Post-Retrieval Processing and Reward Model (PRP-RM). Given a problem \( P \) and its retrieved context \( \mathcal{R}_p \) from the Knowledge Graph (KG), PRP-RM refines it into \( \mathcal{R'}_p \). The Reasoner LLM generates step \( S_1 \) based on \( \mathcal{R'}_p \), followed by iterative retrieval and refinement (\( \mathcal{R}_t \to \mathcal{R'}_t \)) for each step \( S_t \). Correctness is assessed using \( I = \) "Is this step correct?" to compute \( \operatorname{Score}(S_t) \), while completion is checked via \( I_E = \) "Has a final answer been reached?" to compute \( \operatorname{End}(S_t) \). The process continues until \( \operatorname{End}(S_t) \) surpasses a threshold or a predefined inference depth is reached.}
\label{fig:prprm}
\end{center}
\vskip -0.1in
\end{figure}

\noindent\textbf{Iterative Refinement and Verification.}
Inspired by generative reward models \cite{zhang2024generative, mahan2024generative}, we integrate retrieval refinement as a form of CoT reasoning before scoring each step. To ensure consistency in multi-step reasoning, we employ an iterative retrieval refinement and scoring mechanism, as illustrated in Figure~\ref{fig:prprm}.

\begin{algorithm}[t]
\caption{KG-RAR for Step Retrieval}
\setcounter{AlgoLine}{0}
\KwIn{Current step \(S\) and retrieved problems \(\{P_1, \ldots, P_k\}\)}
\KwOut{Relevant step \(S^*\) and its context subgraph}
\BlankLine
Initialize step collection \(V_S \gets \bigcup_{i=1}^{k} \text{Steps}(P_i)\)\;
\BlankLine
\ForEach{\(S_i \in V_S\)}{
  Compute semantic similarity \(\text{Sim}_{\text{semantic}}(S, S_i)\)\;
}
\BlankLine
\(S^* \gets \arg\max_{S_i \in V_S} \text{Sim}_{\text{semantic}}(S, S_i)\)\;
\BlankLine
Construct context subgraph via \(\text{BFS}(S^*)\)\;
\BlankLine
\Return \(S^*\), \(\text{subgraph}(S^*)\)\;
\end{algorithm}

For a reasoning step \( S_t \), the iterative refinement history is:
\[
H_{t} = \{P + \mathcal{R}_p, \mathcal{R}'_p, S_1 + \mathcal{R}_1, \mathcal{R}'_1, \dots, S_{t} + \mathcal{R}_{t}, \mathcal{R}'_{t}\}.
\]
The refined retrieval context is generated recursively:
\[
\mathcal{R}'_t = \operatorname{LLM}_{\text{refine}}(H_{t-1}, S_t + \mathcal{R}_t).
\]
The correctness and end-of-reasoning probabilities are:
\[
\operatorname{Score}(S_t) = \frac{\exp(p(\text{Yes} | H_t, I))}{\exp(p(\text{Yes} | H_t, I)) + \exp(p(\text{No} | H_t, I))},
\]
\[
\operatorname{End}(S_t) = \frac{\exp(p(\text{Yes} | H_t, I_E))}{\exp(p(\text{Yes} | H_t, I_E)) + \exp(p(\text{No} | H_t, I_E))}.
\]

This process iterates until \( \operatorname{End}(S_t) > \theta \), signaling completion.

\begin{table*}[t]
    \centering
    \caption{Performance evaluation across different levels of the Math500 dataset using various models and methods.}
    \vspace{-1.5mm}
    \label{table:math500}
    \resizebox{\linewidth}{!}{%
    {\renewcommand{\arraystretch}{0.9}
    \begin{tabular}{llcccccccccccc}
    \toprule
    \multicolumn{2}{c}{\textbf{Dataset: Math500}} & 
    \multicolumn{2}{c}{\shortstack{\textbf{Level 1} \\ \footnotesize (+9.09\%)}} & 
    \multicolumn{2}{c}{\shortstack{\textbf{Level 2} \\ \footnotesize (+5.38\%)}} & 
    \multicolumn{2}{c}{\shortstack{\textbf{Level 3} \\ \footnotesize (+8.90\%)}} & 
    \multicolumn{2}{c}{\shortstack{\textbf{Level 4} \\ \footnotesize (+7.61\%)}} & 
    \multicolumn{2}{c}{\shortstack{\textbf{Level 5} \\ \footnotesize (+16.43\%)}} & 
    \multicolumn{2}{c}{\shortstack{\textbf{Overall} \\ \footnotesize (+8.95\%)}} \\ 
    \cmidrule(lr){3-4} \cmidrule(lr){5-6} \cmidrule(lr){7-8} \cmidrule(lr){9-10} \cmidrule(lr){11-12} \cmidrule(lr){13-14}
    \textbf{Model} & \textbf{Method} & 
    \textbf{Maj} & \textbf{Last} & 
    \textbf{Maj} & \textbf{Last} & 
    \textbf{Maj} & \textbf{Last} & 
    \textbf{Maj} & \textbf{Last} & 
    \textbf{Maj} & \textbf{Last} & 
    \textbf{Maj} & \textbf{Last} \\ 
    \midrule
    \multirow{2}{*}{\shortstack{Llama-3.1-8B \\ \footnotesize (+15.22\%)}} 
        & CoT-prompting & 80.6 & 80.6 & 74.1 & 74.1 & 59.4 & 59.4 & 46.4 & 46.4 & 27.4 & \textbf{27.1} & 51.9 & 51.9 \\
        & Step-by-Step KG-RAR & \textbf{88.4} & \textbf{81.4} & \textbf{83.3} & \textbf{82.2} & \textbf{70.5} & \textbf{69.5} & \textbf{53.9} & \textbf{53.9} & \textbf{32.1} & 25.4 & \textbf{59.8} & \textbf{57.0} \\ 
    \midrule
    \multirow{2}{*}{\shortstack{Llama-3.2-3B \\ \footnotesize (+20.73\%)}} 
        & CoT-prompting & 63.6 & 65.1 & 61.9 & 61.9 & 51.1 & 51.1 & 43.2 & 43.2 & 20.4 & 20.4 & 43.9 & 44.0 \\
        & Step-by-Step KG-RAR & \textbf{83.7} & \textbf{79.1} & \textbf{68.9} & \textbf{68.9} & \textbf{61.0} & \textbf{52.4} & \textbf{49.2} & \textbf{47.7} & \textbf{29.9} & \textbf{28.4} & \textbf{53.0} & \textbf{50.0} \\ 
    \midrule
    \multirow{2}{*}{\shortstack{Llama-3.2-1B \\ \footnotesize (-4.02\%)}} 
        & CoT-prompting & 64.3 & 64.3 & \textbf{52.6} & \textbf{52.2} & \textbf{41.6} & \textbf{41.6} & \textbf{25.3} & \textbf{25.3} & 8.0 & 8.2 & \textbf{32.3} & \textbf{32.3} \\
        & Step-by-Step KG-RAR & \textbf{72.1} & \textbf{72.1} & 50.0 & 50.0 & 40.0 & 40.0 & 18.0 & 19.5 & \textbf{10.4} & \textbf{13.4} & 31.0 & 32.2 \\ 
    \midrule
    \multirow{2}{*}{\shortstack{Qwen2.5-7B \\ \footnotesize (+2.91\%)}} 
        & CoT-prompting & 95.3 & \textbf{95.3} & 88.9 & 88.9 & 86.7 & 86.3 & 77.3 & 77.1 & 50.0 & 49.8 & 75.6 & 75.4 \\
        & Step-by-Step KG-RAR & 95.3 & 93.0 & \textbf{90.0} & \textbf{90.0} & \textbf{87.6} & \textbf{88.6} & \textbf{79.7} & \textbf{79.7} & \textbf{54.5} & \textbf{56.7} & \textbf{77.8} & \textbf{78.4} \\ 
    \midrule
    \multirow{2}{*}{\shortstack{Qwen2.5-3B \\ \footnotesize (+3.13\%)}} 
        & CoT-prompting & 93.0 & 93.0 & \textbf{85.2} & 85.2 & 81.0 & \textbf{80.6} & 62.5 & 62.5 & 40.1 & \textbf{39.3} & 67.1 & \textbf{66.8} \\
        & Step-by-Step KG-RAR & \textbf{95.3} & \textbf{95.3} & 84.4 & \textbf{85.6} & \textbf{83.8} & 77.1 & \textbf{64.1} & \textbf{64.1} & \textbf{44.0} & 38.1 & \textbf{69.2} & 66.4 \\ 
    \midrule
    \multirow{2}{*}{\shortstack{Qwen2.5-1.5B \\ \footnotesize (-5.12\%)}} 
        & CoT-prompting & 88.4 & 88.4 & 78.5 & \textbf{77.4} & \textbf{71.4} & \textbf{68.9} & \textbf{49.2} & \textbf{49.5} & \textbf{34.6} & \textbf{34.3} & \textbf{58.6} & \textbf{57.9} \\
        & Step-by-Step KG-RAR & \textbf{97.7} & \textbf{93.0} & \textbf{78.9} & 75.6 & 66.7 & 67.6 & 48.4 & 44.5 & 24.6 & 23.1 & 55.6 & 53.4 \\ 
    \bottomrule
    \end{tabular}}%
    }
\end{table*}

\noindent\textbf{Role-Based System Prompting.}
Inspired by agent-based reasoning frameworks \cite{chan2023chateval, talebirad2023multi, zhao2024expel}, we introduce role-based system prompting to further optimize our PRP-RM. In this approach, we define three distinct personas to enhance the reasoning process. The \textbf{Responsible Teacher} \cite{ning2024can} processes retrieved knowledge into structured guidance and evaluates the correctness of each step. The \textbf{Socratic Teacher} \cite{chang2023prompting}, rather than pr oviding direct guidance, reformulates the retrieved content into heuristic questions, encouraging self-reflection. Finally, the \textbf{Critical Teacher} \cite{ke2023critiquellm} acts as a critical evaluator, diagnosing reasoning errors before assigning a score.
Each role focuses on different aspects of post-retrieval processing, improving robustness and interpretability.

	\section{Experiments}
	\label{sec: experiments}

    In this section, we evaluate the effectiveness of our proposed Step-by-Step KG-RAR and PRP-RM methods by comparing them with Chain-of-Thought (CoT) prompting \cite{wei2022chain} and finetuned reward models \cite{cobbe2021gsm8k, lightman2023let}. Additionally, we perform ablation studies to examine the impact of individual components. 
    
	\subsection{Experimental Setup}
	Following prior works \cite{wang2022self, lightman2023let, wang2024openr}, we evaluate on \textbf{Math500} \cite{hendrycksmath2021} and \textbf{GSM8K} \cite{cobbe2021gsm8k}, using \textbf{Accuracy (\%)} as the primary metric. Experiments focus on instruction-tuned \textbf{Llama3} \cite{dubey2024llama} and \textbf{Qwen2.5} \cite{yang2024qwen2}, with \textbf{Best-of-N} \cite{charniak2005coarse, cobbe2021gsm8k, lightman2023let} search (\( n = 8 \) for Math500, \( n = 4 \) for GSM8K). We employ \textbf{Majority Vote} \cite{wang2022self} for self-consistency and \textbf{Last Vote} \cite{snell2024scaling} for benchmarking reward models. To evaluate \textbf{PRP-RM}, we compare against fine-tuned reward models: \texttt{Math-Shepherd-PRM-7B} \cite{wang2024math}, \texttt{RLHFlow-ORM-Deepseek-8B} and \texttt{RLHFlow-PRM-Deepseek-8B} \cite{xiong2024rlhflowmath}. For \textbf{Step-by-Step KG-RAR}, we set step depth to \( 8 \) and padding to \( 4 \). The \textbf{Socratic Teacher} role is used in \textbf{PRP-RM} to minimize direct solving. Both the Reasoner LLM and PRP-RM remain consistent for fair comparison.

\begin{figure}[t]
\begin{center}
\centerline{\includegraphics[width=\columnwidth]{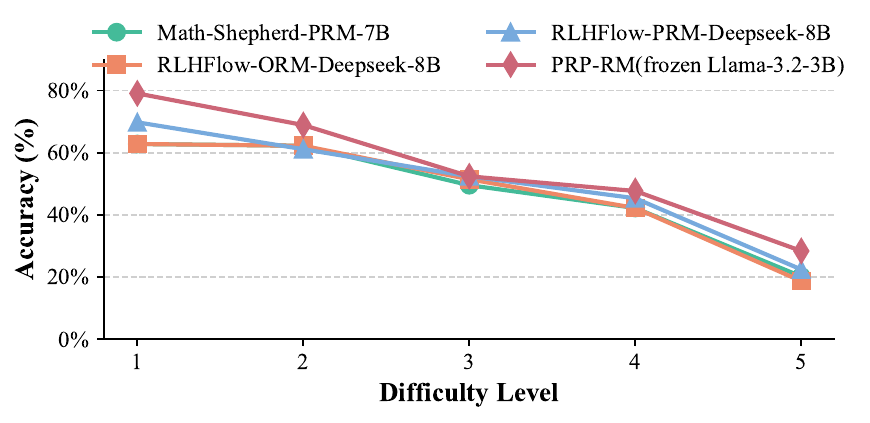}}
\vspace{-2mm}
\caption{Comparison of reward models under Last@8.}
\label{fig:orm-prm-prprm}
\end{center}
    \vspace{-2em}
\end{figure}

	\subsection{Comparative Experimental Results}
	\label{main results}

\begin{table}[t]
\centering
\caption{Evaluation results on the GSM8K dataset.}
\vspace{-1mm}
\label{table:gsm8k}
\resizebox{\columnwidth}{!}{ 
\renewcommand{\arraystretch}{1.0} 
\begin{tabular}{llcc}
\toprule 
\textbf{Model} & \textbf{Method} & \textbf{Maj@4} & \textbf{Last@4} \\ 
\midrule
\multirow{2}{*}{\shortstack{Llama-3.1-8B \\ \footnotesize (+8.68\%)}} 
    & CoT-prompting & 81.8 & 82.0 \\
    & Step-by-Step KG-RAR & \textbf{88.9} & \textbf{88.0} \\ 
\midrule
\multirow{2}{*}{\shortstack{Qwen-2.5-7B \\ \footnotesize (+1.09\%)}} 
    & CoT-prompting & 91.6 & 91.1 \\
    & Step-by-Step KG-RAR & \textbf{92.6} & \textbf{93.1} \\ 
\bottomrule
\end{tabular}
}
\end{table}

Table~\ref{table:math500} shows that Step-by-Step KG-RAR consistently outperforms CoT-prompting across all difficulty levels on Math500, with more pronounced improvements in the Llama3 series compared to Qwen2.5, likely due to Qwen2.5's higher baseline accuracy leaving less room for improvement. Performance declines for smaller models like Qwen-1.5B and Llama-1B on harder problems due to increased reasoning inconsistencies. Among models showing improvements, Step-by-Step KG-RAR achieves an average relative accuracy gain of \textbf{8.95\%} on Math500 under Maj@8, while Llama-3.2-8B attains a \textbf{8.68\%} improvement on GSM8K under Maj@4 (Table~\ref{table:gsm8k}). Additionally, PRP-RM achieves comparable performance to ORM and PRM. Figure~\ref{fig:orm-prm-prprm} confirms its effectiveness with Llama-3B on Math500, highlighting its viability as a training-free alternative.

\begin{figure}[t]
\centering
\begin{minipage}[b]{0.48\columnwidth}
    \centering
    \includegraphics[width=\columnwidth]{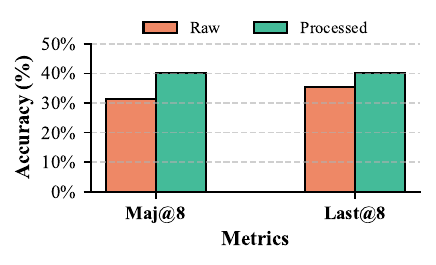}
    
    \caption{Comparison of raw and processed retrieval.}
    \label{fig:raw-processed}
\end{minipage}
\hfill
\begin{minipage}[b]{0.48\columnwidth}
    \centering
    \includegraphics[width=\columnwidth]{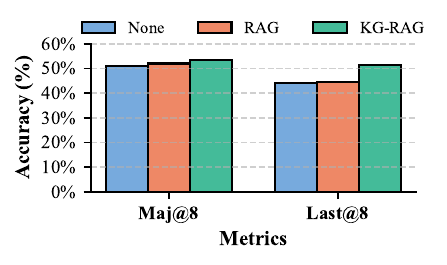}
    \caption{Comparison of various RAG types.}
    \label{fig:none-rag-kgrag}
\end{minipage}
    \vspace{-1.5em}
\end{figure}

\subsection{Ablation Studies}

\noindent\textbf{Effectiveness of Post-Retrieval Processing (PRP).} 
We compare reasoning with refined retrieval from PRP-RM against raw retrieval directly from Knowledge Graphs. Figure~\ref{fig:raw-processed} shows that refining the retrieval context significantly improves performance, with experiments using Llama-3B on Math500 Level 3.

\noindent\textbf{Effectiveness of Knowledge Graphs (KGs).} 
KG-RAR outperforms both no RAG and unstructured RAG (PRM800K) baselines, demonstrating the advantage of structured retrieval (Figure~\ref{fig:none-rag-kgrag}, Qwen-0.5B Reasoner, Qwen-3B PRP-RM, Math500).

\begin{figure}[H]
    \vspace{-0.5em}
\centering
\begin{minipage}[b]{0.48\columnwidth}
    \centering
    \includegraphics[width=\columnwidth]{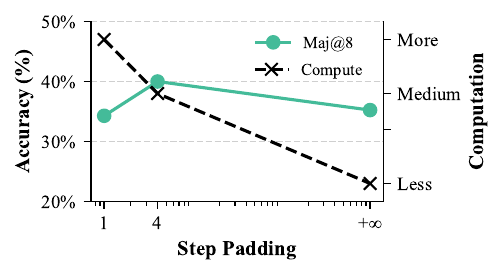}
    \caption{Comparison of step padding settings.}
    \label{fig:steppadding-scaling}
\end{minipage}
\hfill
\begin{minipage}[b]{0.48\columnwidth}
    \centering
    \includegraphics[width=\columnwidth]{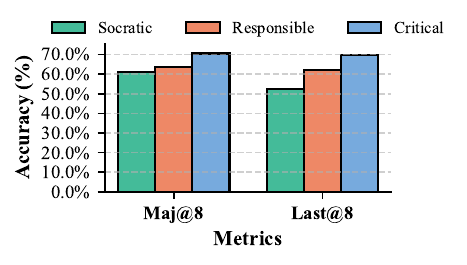}
    \caption{Comparison of PRP-RM roles.}
    \label{fig:socratic-responsible-critical}
\end{minipage}
    \vspace{-1.5em}
\end{figure}

\noindent\textbf{Effectiveness of Step-by-Step RAG.} 
We evaluate step padding at 1, 4, and 1000. Small padding causes inconsistencies, while large padding hinders refinement. Figure~\ref{fig:steppadding-scaling} illustrates this trade-off (Llama-1B, Math500 Level 3).

\begin{figure}[b]
\centering
\begin{minipage}[b]{0.48\columnwidth}
    \centering
    \includegraphics[width=\columnwidth]{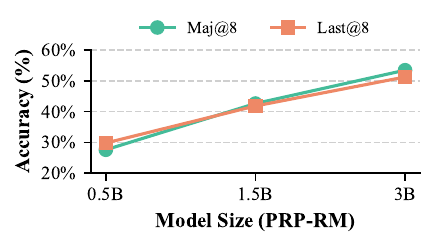}
    \caption{Scaling PRP-RM sizes}
    \label{fig:prprm-scaling}
\end{minipage}
\hfill
\begin{minipage}[b]{0.48\columnwidth}
    \centering
    \includegraphics[width=\columnwidth]{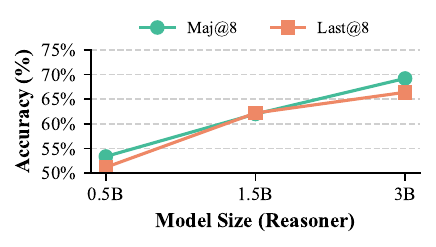}
    \caption{Scaling reasoner LLM sizes}
    \label{fig:reasoner-scaling}
\end{minipage}
\end{figure}

\noindent\textbf{Comparison of PRP-RM Roles.} 
Socratic Teacher minimizes direct problem-solving but sometimes introduces extraneous questions. Figure~\ref{fig:socratic-responsible-critical} shows Critical Teacher performs best among three roles (Llama-3B, Math500).

\noindent\textbf{Scaling Model Size.} 
Scaling trends in Section~\ref{main results} are validated on Math500. Figures~\ref{fig:prprm-scaling} and~\ref{fig:reasoner-scaling} confirm performance gains as both the Reasoner LLM and PRP-RM scale independently.

\begin{figure}[ht]
\vspace{-1em}
\centering
\begin{minipage}[b]{0.48\columnwidth}
    \centering
    \includegraphics[width=\columnwidth]{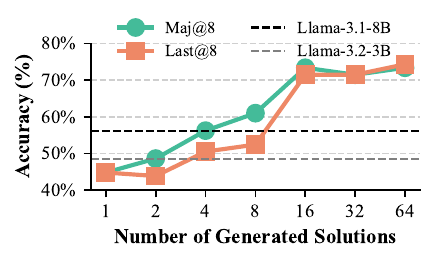}
    \caption{Scaling of the number of solutions.}
    \label{fig:numsolution-accuracy}
\end{minipage}
\hfill
\begin{minipage}[b]{0.48\columnwidth}
    \centering
    \includegraphics[width=\columnwidth]{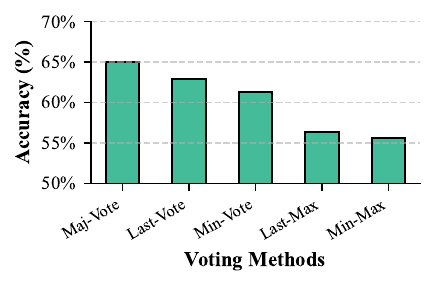}
    \caption{Comparison of various voting methods.}
    \label{fig:voting-ablation}
\end{minipage}
\vspace{-1.5em}
\end{figure}

\noindent\textbf{Scaling of Number of Solutions.} 
We vary the number of generated solutions using Llama-3B on Math500 Level 3. Figure~\ref{fig:numsolution-accuracy} shows accuracy improves incrementally with more solutions, underscoring the benefits of multiple candidates.

\noindent\textbf{Comparison of Voting Methods.} 
We widely evaluate five PRP-RM voting strategies: Majority Vote, Last Vote, Min Vote, Min-Max, and Last-Max \cite{feng2023alphazero, wang2024openr}. Majority Vote and Last Vote outperform others, as extreme-based methods are prone to PRP-RM overconfidence in incorrect solutions (Figure~\ref{fig:voting-ablation}).

\section{Conclusions and Limitations}

    In this paper, we introduce a novel graph-augmented reasoning paradigm that aims to enhance o1-like multi-step reasoning capabilities of frozen LLMs by integrating external KGs. Towards this end, we present \emph{step-by-step knowledge graph based retrieval-augmented reasoning (KG-RAR)}, a novel iterative retrieve-refine-reason framework that strengthens o1-like reasoning, facilitated by an innovative \emph{post-retrieval processing and reward model (PRP-RM)} that refines raw retrievals and assigns step-wise scores to guide reasoning more effectively. Experimental results demonstrate an 8.95\% relative improvement on average over CoT-prompting on Math500, with PRP-RM achieving competitive performance against fine-tuned reward models, yet without the heavy training or fine-tuning costs.

Despite these merits, the proposed approach indeed has some limitations, such as higher computational overhead and potential cases where KG-RAR may introduce unnecessary noise or fail to enhance reasoning. Our future work will focus on optimising the framework by incorporating active learning to dynamically update KGs, improving retrieval efficiency, and exploring broader applications in complex reasoning domains such as scientific discovery and real-world decision-making.

		\ifCLASSOPTIONcaptionsoff
		\newpage
		\fi

		\bibliographystyle{IEEEtran}
		\bibliography{cite.bib}

	\end{document}